\documentclass[runningheads]{llncs}
\usepackage[T1]{fontenc}
\usepackage{color}
\usepackage{graphicx}
\usepackage{cite}
\usepackage{multirow}
\usepackage{booktabs}
\usepackage{subcaption}
\usepackage[colorlinks, linkcolor=black, anchorcolor=black, citecolor=black]{hyperref}
\usepackage{colortbl}

\definecolor{mygray}{gray}{.9}
\definecolor{mycyan}{cmyk}{.3,0,0,0}
\definecolor{mycolor}{rgb}{.90,.90,.98}
\newcolumntype{a}{>{\columncolor{mycolor}}c}

\usepackage{amssymb}
\begin{document}
\title{Learning Discriminative Representation via Metric Learning for Imbalanced Medical Image Classification}
\titlerunning{Learning discriminative representation via metric learning}

\author{
Chenghua Zeng \and 
Huijuan Lu \and
Kanghao Chen \and \\
Ruixuan Wang\thanks{Corresponding author, Email: wangruix5@mail.sysu.edu.cn.} \and
Wei-Shi Zheng
}
\institute{School of Computer Science and Engineering, Sun Yat-sen University, China
}
\maketitle

\begin{abstract}
Data imbalance between common and rare diseases during model training often causes intelligent diagnosis systems to have biased predictions towards common diseases.
The state-of-the-art approaches apply a two-stage learning framework to alleviate the class-imbalance issue, where the first stage focuses on training of a general feature extractor and the second stage focuses on fine-tuning the classifier head for class rebalancing. However, existing two-stage approaches do not consider the fine-grained property between different diseases, often causing the first stage less effective for medical image classification than for natural image classification tasks. In this study, we propose embedding metric learning into the first stage of the two-stage framework specially to help the feature extractor learn to extract more discriminative feature representations. 
Extensive experiments mainly on three medical image datasets show that the proposed approach consistently outperforms existing one-stage and two-stage approaches,
suggesting that metric learning can be used as an effective plug-in component in the two-stage framework for fine-grained class-imbalanced image classification tasks.

\keywords{Class-imbalanced \and metric learning \and two-stage}
\end{abstract}

\section{Introduction}

An ideal intelligent diagnosis system is expected to be able to diagnose both common and rare diseases for specific organs or tissues.
However, while it is relatively easier to collect sufficient training images for common diseases, often much smaller number of images can be collected for rare diseases.
Such data imbalance between classes poses great challenge to learning unbiased classifiers for intelligent diagnosis. To improve the diagnostic performance of the intelligent system especially for those rare diseases,
it is crucial to investigate effective learning strategies which can help the intelligent system successfully learn the features of both common and rare diseases from the imbalanced disease dataset.

Many approaches have been developed to solve the class imbalance issue. Traditional approaches include the re-sampling strategy~\cite{Buda2018ASS} to generate equivalent number of training data for each class,
and the re-weighting strategy to set larger loss weights for training data from infrequent classes~\cite{Cui2019ClassBalanced} or difficult to recognize~\cite{Lin2017Focal}. 
However, these approaches 
often cause classifier over-fitting for the infrequent (minority)  classes due to very limited number of training samples from the infrequent classes. 
To alleviate the over-fitting of the classifier, another group of approaches try to directly improve the generalizability of model, e.g., by transfer learning with a pre-trained classifier backbone using large dataset ImageNet~\cite{Deng2009ImageNet}, or by augmenting the number of training data particularly for infrequent classes with various augmentation techniques like Mixup~\cite{Zhang2018Mixup} and its extensions Remix~\cite{Chou2020Remix} and Balanced-Mixup~\cite{Galdran2021BalancedMixup}.
Additionally, augmentation of infrequent classes in the feature space also helps alleviate the over-fitting issue~\cite{Chu2020FeatureSA}.
Besides the one-stage re-balancing and augmentation strategies, a group of two-stage approaches try to improve the representation ability of the deep neural network, 
first performing representation learning of the feature extractor and then applying re-balancing strategy to the classifier head~\cite{Kang2020DecouplingRA, Cao2019LDAM}.
However, most two-stage approaches focus on the re-balancing strategy for the classifier head, while the feature extractor is simply trained with the plain cross-entropy loss.

In this paper, with the observation that current two-stage approaches do not consistently improve the class-imbalanced medical image classification performance, we propose a new two-stage learning approach mainly by embedding the metric learning to the feature extractor training at the first stage. 
In particular, the feature extractor was trained by minimization of a metric learning loss (e.g., center loss) together with the cross-entropy loss at the first stage. Compared to the plain cross-entropy loss, the inclusion of the metric learning loss can help pull data from the same class together and push data from different classes apart in the feature space. In this way, the feature extractor can capture more discriminative feature representation for fine-grained classes (e.g., different diseases).
Extensive evaluations on multiple medical image datasets support that the proposed approach is effective for class-imbalanced image classification. 

\section{Methodology}

\begin{figure}[t]
    \centering
    \includegraphics[scale=0.48]{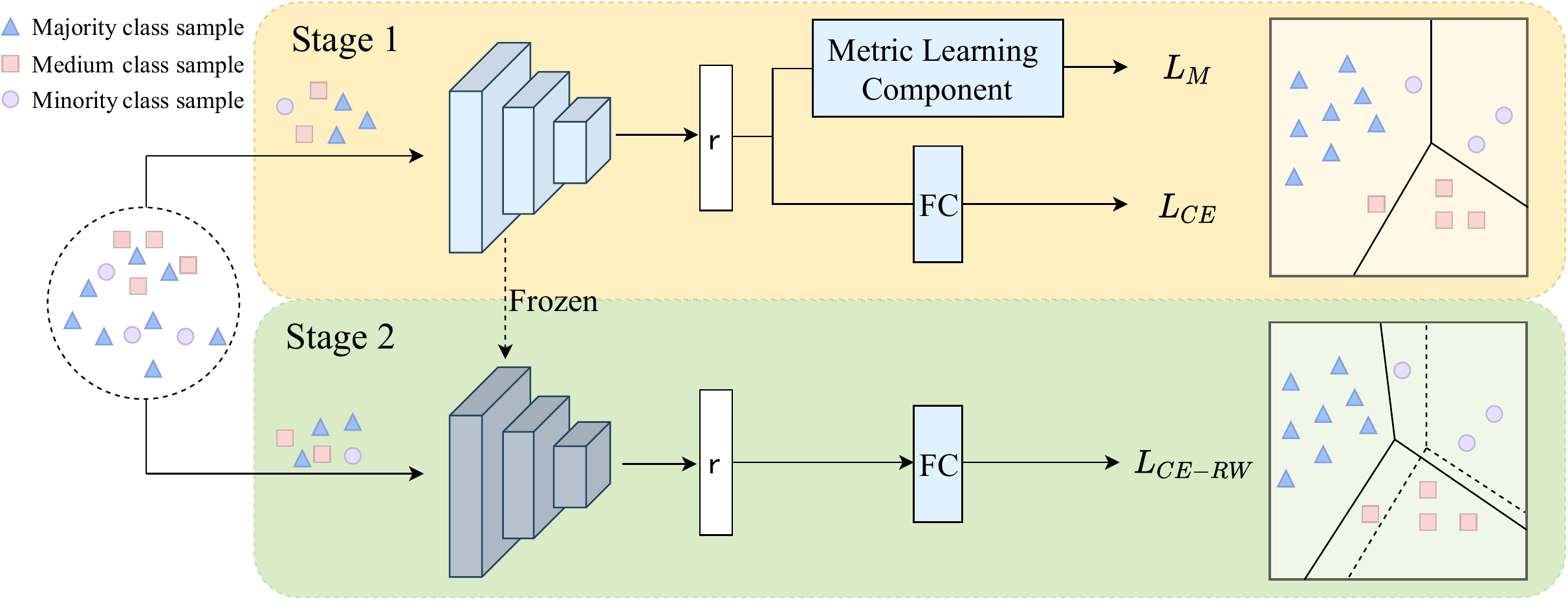}
    \caption{Overview of the proposed two-stage learning approach with the metric learning embedded into the first stage. The metric learning component is not limited to specific metric learning strategy. The feature extractor is frozen and only the classifier head (FC) is fine-tuned with certain rebalancing strategy at the second stage. $\mathbf{r}$: output of the feature extractor. 
    $L_M$: metric learning loss. $L_{CE}$: cross-entropy loss. $L_{CE-RW}$: cross-entropy loss with re-weighting strategy.}
    \label{fig:framework}
\end{figure}

This study focuses on alleviating the effect of data imbalance between classes on medical image classification.
Inspired by the state-of-the-art two-stage framework for class-imbalanced natural image classification and considering the fine-grained property in medical image classification, we propose a new two-stage approach by embedding the metric learning into the first stage (Figure~\ref{fig:framework}).
The approach is not limited to specific metric learning strategy, and therefore various metric learning strategies can be considered in practical applications. 

\subsection{Analysis about classical two-stage framework}
Previous studies on natural image classification have consistently shown that the two-stage learning framework can substantially improve the classification performance when the training data are imbalanced across classes.
At the first stage, a CNN classifier is conventionally trained, e.g., by minimizing the cross-entropy loss without applying any class rebalancing strategy.
At the second stage, the feature extractor part of the trained classifier from the first stage is often fixed, and only the classifier head is re-trained with certain class rebalancing strategy (e.g., class re-weighting).
The widely adopted explanation is that the feature extractor learned at the first stage has a more powerful representation ability than the conventional one-stage class rebalancing strategy, which in turn helps the classifier preserve performance on frequent (majority) classes when trying to improve the performance on the minority classes at the second stage.


While such two-stage learning framework and its variants~\cite{Cao2019LDAM,Kang2020DecouplingRA, Zhou2020BBN, wang2021contrastive} are effective for class-imbalanced natural image classification, we have observed that their effectiveness for class-imbalanced medical image classification is unstable.
One possible reason is that medical image classifications are often fine-grained recognition tasks, i.e., different classes in a medical image classification task are often very similar in appearance and different in certain fine details.
In the case of fine-grained recognition, a feature extractor with general representation power may not be the ideal choice, because quite a large part of features extracted from the feature extractor may be relevant to image backgrounds or class-shared properties rather than class-specific fine-grained details in images.
Therefore, it may be necessary to train a feature extractor which is powerful in extracting more discriminative fine-grained features in order to achieve better performance on class-imbalanced medical image classification. 

\subsection{Metric learning-embedded two-stage learning framework}
With the consideration above, we propose applying metric learning to the training of the feature extractor at the first stage, where the role of metric learning is to make the distribution of each class more compact in the feature space (i.e., the output space of the feature extractor) and meanwhile different classes clearly separated from each other.
By enforcing intra-class compactness and inter-class separation in the feature space, it would be expected that the feature extractor is forced to learn to extract features which are shared (therefore compact) within each class but discriminative (therefore separated) between classes.
Various metric learning strategies can be adopted, e.g., based on the center loss~\cite{wen2016centerloss}, the triplet loss~\cite{Hoffer2015tripletloss}, or the supervised contrastive loss~\cite{Khosla2020SupContrastLoss}.
Note that while the center loss and the triplet loss were directly computed based on 
Euclidean distance in the feature space (i.e., the output space of the feature extractor),
the supervised constrative loss was computed based on the L2-normlized feature representation from the output of an 3-layer MLP following the feature extractor. 
The supervised contastive loss instead of the more widely used unsupervised contastive loss~\cite{Chen2020simclr} was adopted here to help enforce different samples of the same class having similar feature representation and therefore more compact intra-class distribution, where a positive pair corresponds to two different training samples from the same class instead of two augmented versions of one sample. 


It is worth noting that while the solely metric learning loss can be used to train a feature extractor at the first stage, prior studies on metric learning have suggested combining the metric learning loss with the cross-entropy loss for the training of the feature extractor, because it has been consistently observed that the training process becomes more stable and often faster when the cross-entropy loss is minimized together with the metric learning loss.
Without the help of the cross-entropy loss, the change in feature extractor is simply determined by the proximity of a small batch of training data at each iteration.
The potential large difference between different batches of training data could cause large changes (therefore unstable) in feature extractor across training iterations.

\section{Experiment}

\subsection{Experimental setup}\label{section:exp_setup}

\subsubsection{Datasets:} The proposed method was evaluated mainly on three imbalanced medical image datasets, Skin7~\cite{Codella2018SkinLA}, X-ray9, and PathMNIST-LT. 
X-ray9 was selected from the original 14-class ChestX-ray14~\cite{Wang2017ChestXray} by discarding those classes which may appear together with other diseases in the same images.
PathMNIST-LT is the imbalanced version of PathMNIST from MedMNISTv2~\cite{medmnistv2}, by exponentially reducing the number of training samples across classes.
Skin7 and X-ray9 were respectively randomly split to the training set, validation set and test set with the ratio 7.5:0.5:2, while the original validation set and test set of PathMNIST were kept unchanged for PathMNIST-LT.
Table~\ref{tab:dataset} summarizes the main information of the three datasets. It is worth noting that classes are heavily imbalanced and fine-grained from each medical image dataset.

\begin{table}[btp]
\caption{Dataset statistics. $N_{min}$ (or $N_{max}$):  number of training samples from the least (or most) frequent disease. $\rho$:  imbalance ratio, i.e., $\rho={N_{max}}/{N_{min}}$.}
\centering
{%
\begin{tabular}{c|c|c|c|c|c|c}
\toprule
Dataset      & ImageType       & Diseases & ImageSize          & $N_{min}$ & $N_{max}$ & $\rho$ \\ 
\midrule
Skin7        & Dermoscopy      & 7        & $600 \times 450$   & 86        & 5029      & 58.5 \\ 
X-ray9       & X-ray           & 9        & $1024 \times 1024$ & 83        & 3158      & 38.1 \\ 
PathMNIST-LT & Colon Pathology & 9        & $28 \times 28$     & 94        & 9366      & 99.6 \\ 
\bottomrule
\end{tabular}%
}
\label{tab:dataset}
\end{table}

\vspace{-0.3cm}
\subsubsection{Implementation details:} ResNet50 pretrained on ImageNet~\cite{Deng2009ImageNet} was adopted as the default CNN classifier. 
When training a classifier on each dataset, all images were randomly rotated within the angle range [-30\textdegree, 30\textdegree], resized to $300 \times 300$ pixels and then randomly cropped to $224 \times 224$ pixels, followed by a random horizontal flip with probability 0.5.
At the first stage, the stochastic gradient descent(SGD) optimizer with momentum(0.9) and weight decay (0.0005) was adopted, with mini-batch size 32 on Skin7 and 64 on X-ray9 and PathMNIST-LT.
The learning rate was warmed up by linearly increasing to 0.001 in the first 5 epochs and then decayed to 1e-6 over epochs with the  CosineAnnealingLR strategy~\cite{Loshchilov2017SGDR}.
Each model was trained for up to 100 epochs with explicit convergence.
At the second stage, the feature extractor is frozen and only the classifier head was fine-tuned for 10 epochs, with the learning rate initialized as 0.1 and then decayed to 1e-6 over epochs using the CosineAnnealingLR strategy. 
With the help of validation sets, the coefficient for the center loss and the triplet loss  was set to 0.001, and the margin in triplet loss was set 50.
For the contrastive loss,
the coefficient was 1.0 and the temperature was 0.05.
During inference, all images were resized to $300\times300$ pixels and cropped to $224\times224$ pixels at the center.
Due to imbalance in test set for Skin7 and X-ray9,
model performance was measured by the mean class recall (MCR) over all classes, and the MCRs for the majority classes (3 most frequent classes from each dataset) 
and the minority classes (3 most infrequent classes from X-ray9 and PathMNIST-LT; 2 from Skin7) respectively.
For each experiment, 
the mean and standard deviation of MCRs over three runs were reported. 

\begin{table}[!ht]
\centering
\caption{Comparison with various baselines on three medical image datasets. All/Major/Minor: MCR over all/majority/minority classes.  
"$\rightarrow$": from first to second stage.
Subscript values: standard deviation. The best MCR and the second best on each dataset is marked in bold and underline, respectively. }
\resizebox{\linewidth}{!}{
\setlength{\tabcolsep}{1mm}
\begin{tabular}{l|ccccccccc}
\toprule

\multirow{2}{*}{\textbf{Methods}} &
\multicolumn{3}{c}{\textbf{Skin7}} & \multicolumn{3}{c}{\textbf{X-ray9}} & \multicolumn{3}{c}{\textbf{PathMNIST-LT}} \\
\cmidrule(l){2-4}
\cmidrule(l){5-7}
\cmidrule(l){8-10}
~ & All & Major & Minor                  & All & Major & Minor                & All & Major & Minor \\[0.5ex]
\midrule
CE
& 83.12 \tiny{0.89} & 85.09 & 80.06   & 50.20 \tiny{0.75} & 58.49 & 41.82   & 84.46 \tiny{0.11} & 94.34 & 70.21 \\[0.5ex]
CE $\rightarrow$ cRW
& 84.11 \tiny{0.70} & 85.08 & 82.67   & 51.81 \tiny{1.03} & 57.58 & 47.79   & 83.77 \tiny{0.39} & 90.66 & 71.15 \\[0.5ex]
CE $\rightarrow$ cRS
& 82.17 \tiny{0.85} & 85.54 & 80.06   & 51.46 \tiny{0.77} & 57.89 & 46.50   & 82.26 \tiny{0.52} & 88.96 & 70.37 \\[0.5ex]

CE-DRW
& 83.58 \tiny{0.17} & 84.46 & 82.23   & 51.04 \tiny{0.28} & 58.83 & 43.99   & 84.91 \tiny{0.38} & 92.40 & 72.94 \\[0.5ex]
RS
& 81.39 \tiny{0.10} & 85.69 & 77.89   & 47.86 \tiny{0.74} & 59.55 & 38.97   & 84.22 \tiny{0.25} & 94.47 & 68.08 \\[0.5ex]
RW
& 84.35 \tiny{0.94} & 86.36 & 86.86   & 49.44 \tiny{0.28} & 57.59 & 44.07   & 79.88 \tiny{0.17} & 77.76 & 71.88 \\[0.5ex]

LDAM
& 84.37 \tiny{0.32} & 85.50 & 82.23   & 49.07 \tiny{0.51} & 59.49 & 40.20   & 80.86 \tiny{0.32} & 78.10 & 73.31 \\[0.5ex]
LDAM $\rightarrow$ cRW
& 82.36 \tiny{0.41} & 84.35 & 78.79   & 48.19 \tiny{0.38} & 59.71 & 37.78   & 80.65 \tiny{0.35} & 79.52 & 72.18 \\[0.5ex]
LDAM-DRW
& 85.21 \tiny{0.44} & 85.54 & 84.41   & 52.18 \tiny{0.62} & 58.53 & 48.94   & 82.08 \tiny{0.27} & 84.39 & 72.06 \\[0.5ex]

Focal
& 82.66 \tiny{0.73} & 82.93 & 81.78   & 49.86 \tiny{0.56} & 57.34 & 44.48   & 84.62 \tiny{0.33} & 94.42 & 71.86 \\[0.5ex]
Focal $\rightarrow$ cRW
& 83.62 \tiny{0.58} & 83.64 & 86.13   & 51.47 \tiny{0.36} & 54.95 & 51.51   & 80.83 \tiny{0.29} & 85.76 & 69.70 \\[0.5ex]

CB-Focal
& 78.25 \tiny{0.93} & 72.52 & 80.06   & 46.30 \tiny{0.75} & 42.13 & 53.59   & 82.79 \tiny{0.44} & 91.72 & 72.75 \\[0.5ex]
CB-Focal $\rightarrow$ cRW
& 76.95 \tiny{0.79} & 75.32 & 80.06   & 46.42 \tiny{0.53} & 42.04 & 55.16   & 79.32 \tiny{0.31} & 88.92 & 68.63 \\[0.5ex]

Mixup
& 81.65 \tiny{0.28} & 85.39 & 81.87   & 49.54 \tiny{0.33} & 59.47 & 42.20   & 84.26 \tiny{0.14} & 96.73 & 68.07 \\[0.5ex]
Mixup $\rightarrow$ cRW
& 83.37 \tiny{0.34} & 86.70 & 81.78   & 50.55 \tiny{0.41} & 59.37 & 45.17   & 83.01 \tiny{0.21} & 91.35 & 68.94 \\[0.5ex]

CutMix
& 82.02 \tiny{0.46} & 86.03 & 79.61   & 50.97 \tiny{0.54} & 59.34 & 43.41   & 84.63 \tiny{0.13} & 95.98 & 67.38 \\[0.5ex]
CutMix $\rightarrow$ cRW
& 84.68 \tiny{0.34} & 87.30 & 83.51   & 52.87 \tiny{0.27} & 55.48 & 53.72   & 85.03 \tiny{0.17} & 84.85 & 68.61 \\[0.5ex]

BBN
& 73.88 \tiny{0.35} & 79.16 & 80.51   & 48.11 \tiny{0.42} & 38.80 & 61.53   & 84.22 \tiny{0.18} & 94.28 & 70.79 \\[0.5ex]

\midrule






\rowcolor{mycolor}
\textbf{CE+CT} $\rightarrow$ \textbf{cRW}
& 85.24 \tiny{0.58} & 86.10 & 83.51
& 52.71 \tiny{0.35} & 56.43 & 53.29
& 85.81 \tiny{0.36} & 94.32 & 74.03 \\[0.5ex]

\rowcolor{mycolor}
\textbf{CE+TP} $\rightarrow$ \textbf{cRW}
& \textbf{86.50} \tiny{0.34} & 84.79 & 86.13
& \textbf{52.93} \tiny{0.26} & 57.90 & 48.83
& \underline{86.50} \tiny{0.42} & 94.85 & 74.12 \\[0.5ex]

\rowcolor{mycolor}
\textbf{CE+SC} $\rightarrow$ \textbf{cRW} 
& \underline{86.43} \tiny{0.44} & 85.73 & 91.75   
& \underline{52.77} \tiny{0.33} & 57.80 & 49.49   
& \textbf{87.45}    \tiny{0.25} & 96.01 & 76.81 \\[0.5ex] 

\bottomrule
\end{tabular}
}
\label{tab:effectiveness}
\end{table}

\subsection{Comparisons with existing approaches}
The proposed two-stage approach was compared with one-stage class rebalancing approaches including re-sampling(RS), re-weighting(RW), LDAM~\cite{Cao2019LDAM}, LDAM with deferred re-weighting strategy (LDAM-DRW)~\cite{Cao2019LDAM}, class-balanced focal loss (CB-Focal)~\cite{Cui2019ClassBalanced, Lin2017Focal}, BBN~\cite{Zhou2020BBN},
Mixup~\cite{Zhang2018Mixup}, CutMix~\cite{Yun2019CutMix}, and two-stage approaches including classifier re-training by re-weighting (cRW) or re-sampling (cRS)~\cite{Kang2020DecouplingRA}, and the combination of classifier re-training and one-stage approaches listed above. 
The center loss (CT), triplet loss (TP) and supervised contrastive loss (SC) were adopted respectively in our approach.
On all three medical image datasets, Table~\ref{tab:effectiveness} shows that the classifiers trained with our approach (last 3 rows) consistently overall outperform existing one-stage and two-stage approaches.
In particular, the improvements by our approach are more significant particularly on minority classes, 
 with little performance drop on majority classes. 

Note that although existing two-stage approaches were reported to outperform conventional one-stage approaches on various natural image classification tasks (e.g., CIFAR-LT~\cite{Cao2019LDAM}, ImageNet-LT~\cite{Liu2019ImageNetLT}, iNaturalist~\cite{Horn2018iNaturalist}), the results here on medical image classifications did not consistently support such a conclusion (e.g., see CE, Focal, CB-Focal, CutMix, Mixup versus their two-stage versions in Table~\ref{tab:effectiveness}).
These inconsistent findings support that the general feature extractor training from existing two-stage approaches may not work well for fine-grained recognition tasks like medical image classification here.
In contrast, the proposed metric learning embedded feature learning would more likely help the feature extractor learn to extract discriminative features.

\subsection{Ablation and sensitivity study}
An ablation study was performed to further confirm the effectiveness of our two-stage approach.
Table~\ref{tab:ablation} shows that, when respectively dropping each component (CE, SC, cRW; rows 2-4) in training, the classifier performance becomes worse on both overall classes (`All') and  minority classes (`Minor'). In particular, by including the embedded metric learning to the two-stage framework (third row vs. last row), the classification performance was boosted significantly, directly approving the role of metric learning for class-imbalanced image classification. 


\begin{table}[t]
\setlength{\tabcolsep}{2mm}
\caption{Ablation study on Skin7 and X-ray9.}
    \centering
    \begin{tabular}{ccc|cccccc}
    \toprule
        
        \multirow{2}{*}{CE} & \multirow{2}{*}{SC} & \multirow{2}{*}{cRW} & \multicolumn{3}{c}{Skin7} & \multicolumn{3}{c}{X-ray9} \\
        \cmidrule(l){4-6}
        \cmidrule(l){7-9}
         ~         & ~          & ~          & All   & Major & Minor  & All   & Major & Minor \\[0.5ex]
        \midrule
        \checkmark & -          & -          & 83.12 & 85.09 & 80.06  & 50.20 & 58.49 & 41.82 \\[0.5ex]
        \checkmark & \checkmark & -          & 83.56 & 86.74 & 80.06  & 50.97 & 60.49 & 42.64 \\[0.5ex]
        \checkmark & -          & \checkmark & 84.11 & 85.08 & 82.67  & 51.81 & 57.58 & 47.79 \\[0.5ex]
        -          & \checkmark & \checkmark & 84.36 & 84.12 & 86.13  & 49.71 & 54.05 & 40.74 \\[0.5ex]
        \checkmark & \checkmark & \checkmark & 86.43 & 85.73 & 91.75  & 52.77 & 57.80 & 49.49 \\[0.5ex]
        \bottomrule
    \end{tabular}
    \label{tab:ablation}
\end{table}

\begin{figure}[t]
    \centering
    \includegraphics[height =0.25\textwidth,width=0.32\textwidth]{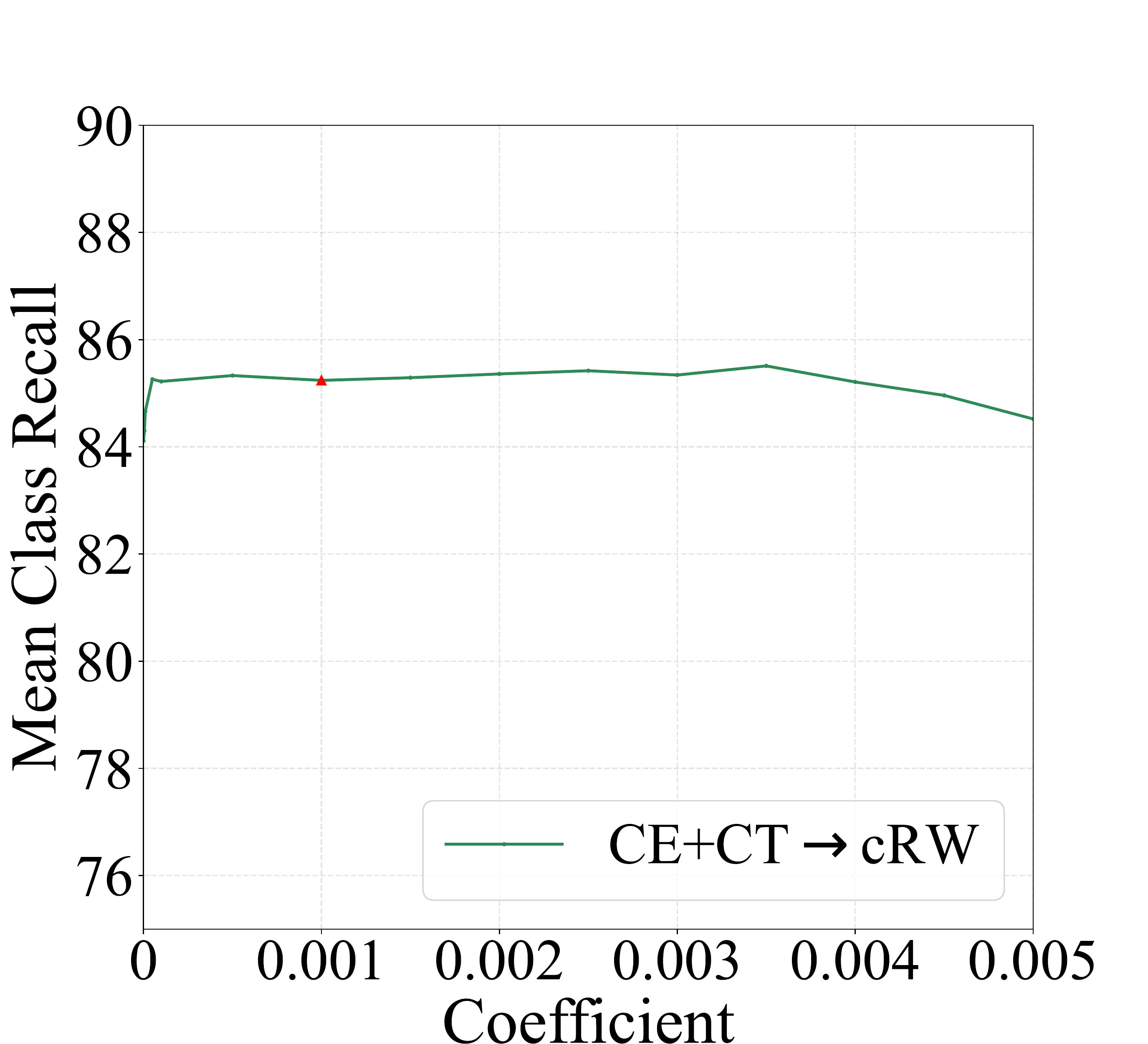}
    \includegraphics[height =0.25\textwidth,width=0.32\textwidth]{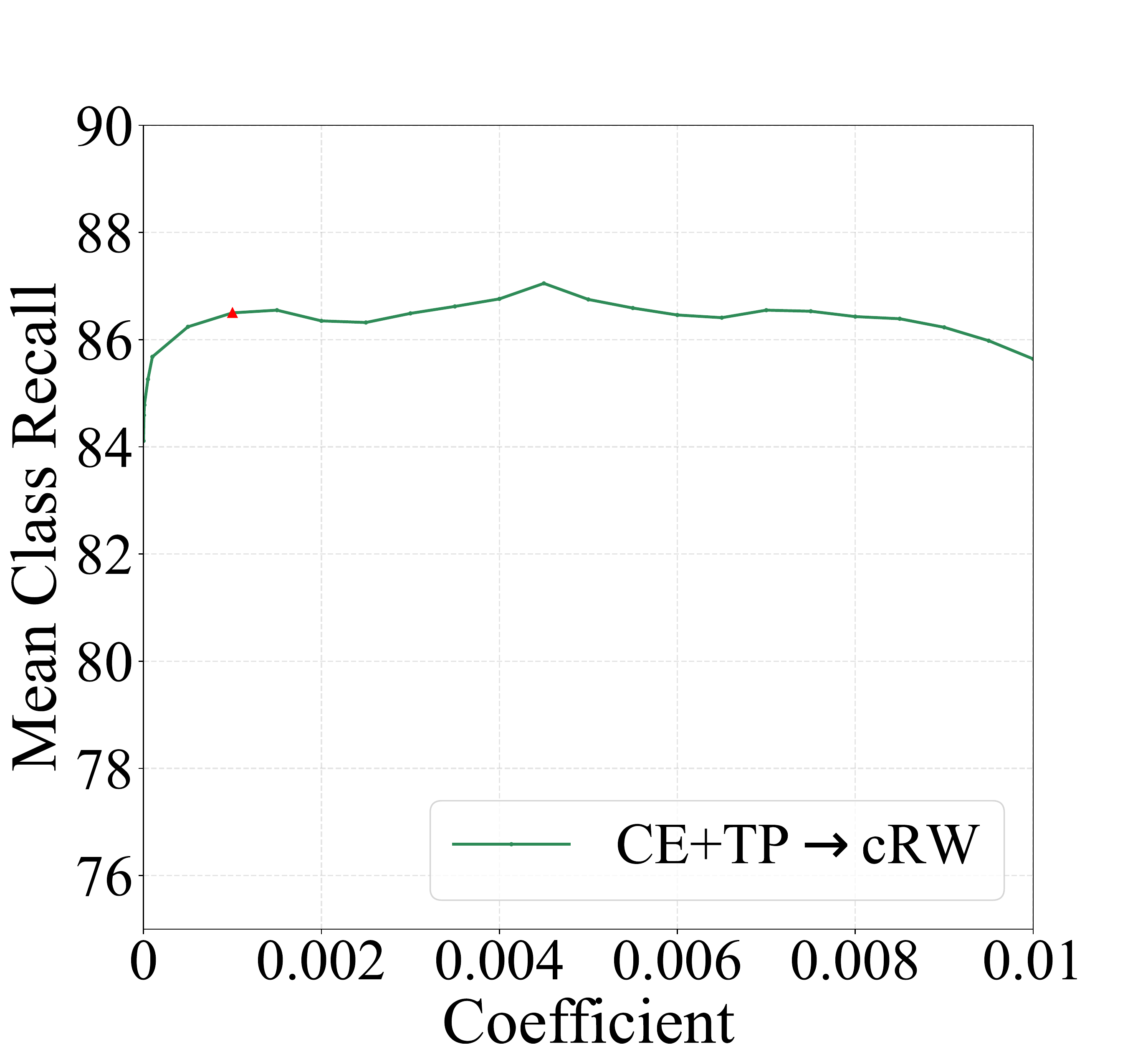}
    \includegraphics[height =0.25\textwidth,width=0.32\textwidth]{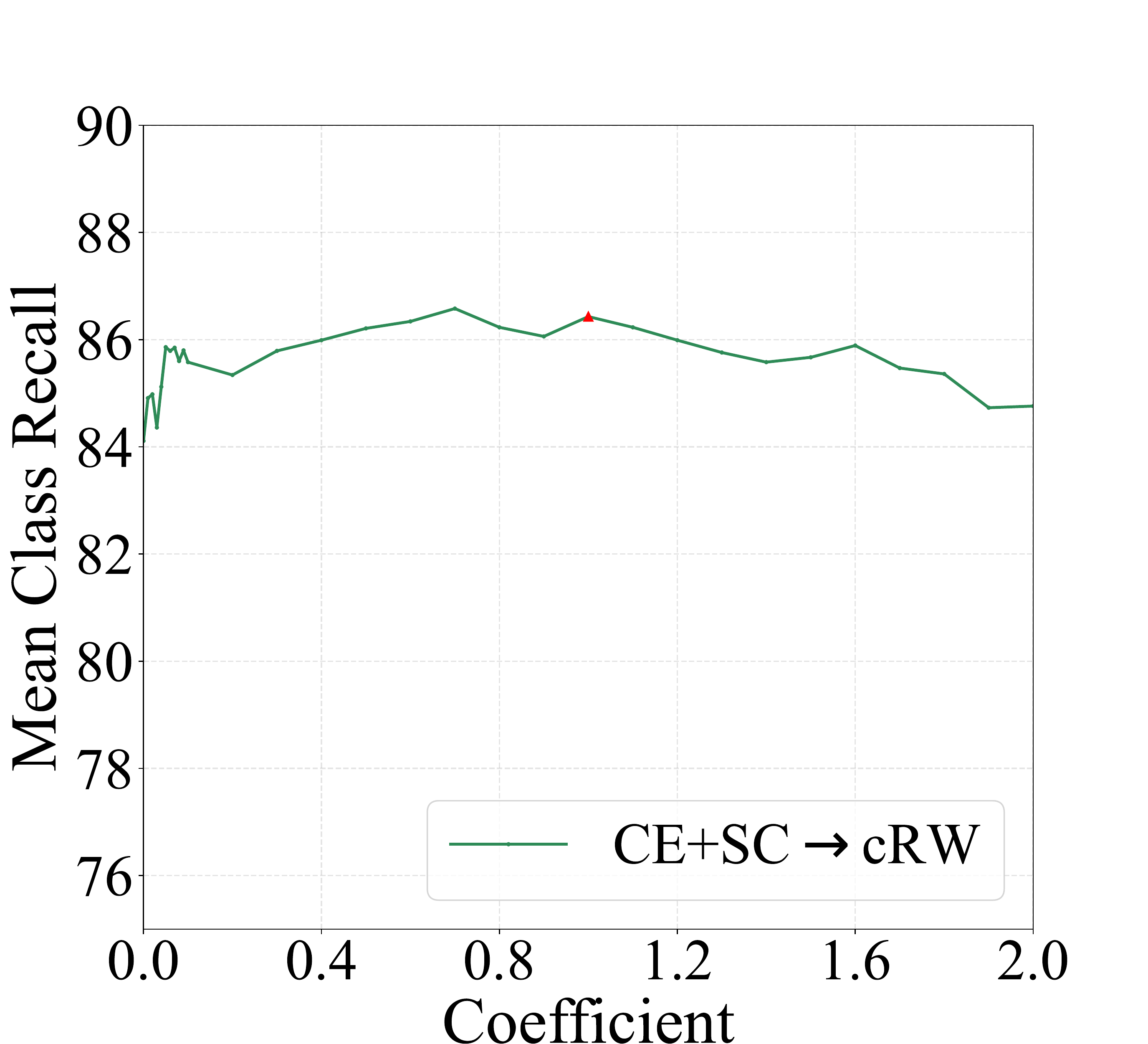}
    
    
    
    
    \caption{Sensitivity study of coefficients for metric learning loss (green curves). From left to right: classification performance when using the center loss, the triplet loss, and the supervised contrastive loss respectively. 
    The default value adopted in experiments is marked as red triangle.}
    \label{fig:sensitive_lambda}
\end{figure}

Another set of studies were performed to evaluate the sensitivity of the approach to the coefficient of the metric learning loss.
From Figure~\ref{fig:sensitive_lambda}, it can be observed that the classification performance is relatively stable in certain range of coefficient values for each metric learning loss term, i.e., [0.00005, 0.004] for center loss, [0.00005, 0.01] for triplet loss, [0.04, 1.8] for supervised contrastive loss. The working coefficient values for the center loss and triplet loss are often smaller than for the supervised contrastive loss because the center loss and the triplet loss are based on Euclidean distance between (raw) feature vectors in the feature space, which often causes large loss values compared to the supervised contrastive loss which is based on the L2 normalized feature vectors.

\subsection{Generalizability of the proposed approach}
The proposed framework is not limited to specific CNN backbones for the feature extractor.
Besides the default ResNet50 backbone, two other CNN backbones were also adopted to confirm the generalizability of the proposed framework.
As Table~\ref{tab:backbone} demonstrates, with each  backbone (DenseNet121~\cite{Huang2017DenseNet} and MobileNetV2~\cite{Sandler2018MobileNetV2}) the classifier trained with the proposed approach (last 3 columns) always outperforms the strongest baselines (first 3 columns) when measured over all classes, and performs better mostly when measured over the minority classes.


\begin{table}[t]
    \setlength{\tabcolsep}{1.5mm}
    \caption{Performance on Skin7 with DenseNet121 and MobileNetV2 backbones. The MCRs of `All'/`Minor' classes were reported for each method.}
    \centering
    \resizebox{\linewidth}{!}{
    \begin{tabular}{l|ccc|ccc}
    \toprule
         \multirow{2}{*}{Method} & CE & \multirow{2}{*}{LDAM-DRW} & CutMix & CE+CT & CE+TP & CE+SC \\
         ~ & $\rightarrow$cRW &  & $\rightarrow$cRW & $\rightarrow$cRW & $\rightarrow$cRW & $\rightarrow$cRW \\
         \midrule
         DenseNet121 & 84.65/87.86 & 84.79/85.68 & 83.71/86.13 & 85.37/86.93 & \textbf{86.30}/90.21 & \underline{85.93}/87.96 \\
         MobileNetV2 & 83.66/83.51 & 84.09/80.06 & 82.24/80.16 & 84.93/87.86 & \underline{85.96}/89.94 & \textbf{86.26}/86.69 \\
        \bottomrule
    \end{tabular}
    }
    \label{tab:backbone}
\end{table}

Also, considering that most two-stage approaches were developed and evaluated on natural image datasets, an additional evaluation was performed on 
two natural fine-grained image datasets Flower~\cite{Nilsback2008Flower} and Stanford-Dogs~\cite{khosla2011StanfordDogs}.
Both datasets were pre-processed to become class-imbalanced (re-named as Flower-LT and Stanford Dogs-LT) with imbalance ratio $\rho=10$, as for the PathMNIST-LT. 
All classifiers were trained from scratch considering certain classes in each dataset are overlapped with those of ImageNet.
From Table~\ref{tab:flowers}, it can be observed that, 
all the strong baselines were outperformed by the proposed two-stage approach particularly when the center loss (CT) or supervised contrastive loss (SC) was used in training. This further confirms that the metric learning at the first stage is effective in improving the class-imbalanced image classification for the fine-grained recognition tasks.

\begin{table}[t]
\setlength{\tabcolsep}{2mm}
\centering
\caption{Performance on two natural image datasets. 
}
\resizebox{\linewidth}{!}{
\begin{tabular}{l|cccccccc}
\toprule
\multirow{2}{*}{Method} & \multicolumn{4}{c}{Flower-LT} & \multicolumn{4}{c}{Stanford Dogs-LT} \\
\cmidrule(l){2-5}
\cmidrule(l){6-9}
~                        & All  & Major & Medium & Minor  & All & Major  & Medium & Minor  \\
\midrule
CE                       & 88.68 & 92.82 & 90.85 & 82.36  & 28.47 & 42.47 & 28.71 & 14.24  \\
CE $\rightarrow$ cRW     & 89.60 & 92.55 & 90.80 & 85.44  & 29.04 & 40.37 & 29.97 & 16.77  \\
LDAM-DRW                 & 90.43 & 92.39 & 90.27 & 88.63  & 24.75 & 37.30 & 27.26 & 12.83  \\
CutMix $\rightarrow$ cRW & 90.16 & 93.66 & 89.52 & 87.30  & 26.20 & 38.46 & 26.71 & 13.42  \\
\midrule
CE+CT $\rightarrow$ cRW  & \underline{90.94} & 92.29 & 90.15 & 89.60  & \textbf{34.71} & 46.53 & 36.42 & 21.19  \\
CE+TP $\rightarrow$ cRW  & 90.78 & 93.69 & 91.50 & 87.45  & 32.47 & 44.46 & 34.06 & 18.89  \\
CE+SC $\rightarrow$ cRW  & \textbf{90.99} & 93.44 & 90.09 & 89.45  & \underline{34.49} & 47.38 & 35.77 & 20.32  \\
\bottomrule
\end{tabular}
}
\label{tab:flowers}
\end{table}

\section{Conclusion}
In this work, we propose a new two-stage learning approach by embedding the metric learning into the first stage to help train a  feature extractor whose output is more discriminative and therefore suitable for  fine-grained recognition tasks as in medical image classification.
Experiments on three class-imbalanced medical image datasets support that the proposed two-stage approach consistently outperforms all existing class rebalancing methods and their extended two-stage versions. 
The effectiveness of the proposed approach on multiple metric learning losses, model backbones, and natural image datasets further support that the metric learning can be used as a plug-in component to improve class-imbalanced image classification performance in multiple data domains.


\bibliographystyle{splncs04.bst}
\bibliography{main}
\end{document}